\pdfoutput=1

\documentclass[11pt]{article}

\usepackage{EACL2023}

\usepackage{times}
\usepackage{latexsym}

\usepackage[T1]{fontenc}

\usepackage[utf8]{inputenc}

\usepackage{microtype}

\usepackage{inconsolata}

\usepackage{times}
\usepackage{latexsym}
\usepackage{color,soul}
\usepackage{array,multirow,graphicx}
\usepackage{makecell}

\usepackage{amsmath}

\usepackage{microtype}
\usepackage{cleveref}

\usepackage{bbm}

\usepackage{booktabs}

\usepackage{xcolor}
\usepackage[linesnumbered,ruled,vlined]{algorithm2e}

\SetCommentSty{mycommfont}

\SetKwInput{KwInput}{Input}                
\SetKwInput{KwOutput}{Output}              

\usepackage[colorinlistoftodos,prependcaption,textsize=tiny,disable]{todonotes}

\newcommand{\di}[1]{\todo[color=orange!20] {di: #1}}

\title{Selective In-Context Data Augmentation for Intent Detection using Pointwise V-Information}

\author{
Yen-Ting Lin\thanks{~~ Work done during internship at Amazon Alexa AI} \quad Alexandros Papangelis$^\dag$\quad Seokhwan Kim$^\dag$\quad Sungjin Lee$^\dag$ \\
\textbf{Devamanyu Hazarika$^\dag$ \quad Mahdi Namazifar$^\dag$ \quad Di Jin$^\dag$ \quad Yang Liu$^\dag$ \quad Dilek Hakkani-Tur$^\dag$}\\
  $^\star$National Taiwan University \\
  $^\dag$Amazon Alexa AI \\
  \small\texttt{ytl@ieee.org} \quad  \texttt{\{papangea,seokhwk,sungjinl,dvhaz,mahdinam,djinamzn,yangliud,hakkanit\}@amazon.com} \\
}

\begin{document}
\maketitle
\begin{abstract}
This work focuses on in-context data augmentation for intent detection. Having found that augmentation via in-context prompting of large pre-trained language models (PLMs) alone does not improve performance, we introduce a novel approach based on PLMs and pointwise V-information (PVI), a metric that can measure the usefulness of a datapoint for training a model.
Our method first fine-tunes a PLM on a small seed of training data and then synthesizes new datapoints -- utterances that correspond to given intents.
It then employs intent-aware filtering, based on PVI, to remove datapoints that are not helpful to the downstream intent classifier. Our method is thus able to leverage the expressive power of large language models to produce diverse training data. 
Empirical results demonstrate that our method can produce synthetic training data that achieve state-of-the-art performance on three challenging intent detection datasets under few-shot settings (1.28\% absolute improvement in 5-shot and 1.18\% absolute in 10-shot, on average) and perform on par with the state-of-the-art in full-shot settings (within 0.01\% absolute, on average).
\end{abstract}

\section{Introduction}
Intent detection, defined as the identification of a user's intent given an utterance, is a fundamental element in task-oriented dialogue systems, usually occurring within the Natural Language Understanding (NLU) component.
One of the practical challenges of training and deploying NLU modules is data scarcity, due to various reasons, such as under-represented languages, privacy and ethical concerns, or simply the cost of collecting and annotating sufficiently large amounts of data for new intents. 
Consequently, accurately identifying intents in limited-resource scenarios has drawn attention from the community \cite[for example]{papangelis-etal-2021-generative,mehri-eric-2021-example,zhang-etal-2021-shot}. 

There are three main families of approaches that address the challenge of limited data for intent detection: data augmentation \cite{DBLP:conf/interspeech/PengZZG21, DBLP:conf/iclr/LiYHLNRYZX21}, focusing on generating high-quality synthetic training and evaluation data; few-shot learning \cite{zhang-etal-2020-discriminative,zhang-etal-2021-shot}, focusing on creating learning algorithms that can cope with limited amounts of data; and transfer learning \cite{DBLP:conf/icassp/NamazifarPTH21}, focusing on learning algorithms that can generalize across domains (therefore not requiring in-domain data). In this work, we follow the data augmentation approach, which is a general method that attempts to augment a human-authored dataset with a large set of synthetically-generated instances.
Most recent work has suggested using Pre-trained Language Models (PLMs) for data augmentations under various setups, e.g., \cite{DBLP:conf/interspeech/PengZZG21}, showing great improvements in performance.
However, simply generating a large number of synthetic data points is not enough; we need to consider the quality of each data point, i.e., how beneficial it would be to the model's performance if that synthetic data point is added to the training set. This is an important issue since the model might learn to overfit to synthetic datapoints (which may be low quality, represent specific use cases, etc.) and thus under-perform on real data.

\di{Shall we provide some concrete examples here to showcase that previous methods may generate low-quality samples that are not relevant to the target labels. Dev: I agree, it will be very strong to put some examples in the appendix and refer to it here.}

In this work, we propose to apply Pointwise $\mathcal{V}$-Information (PVI) \cite{DBLP:conf/icml/EthayarajhCS22} for data augmentation, in a way that leverages a PLM to generate synthetic examples that are relevant and beneficial for training the downstream model, which in our case is an intent classifier. Our contributions are as follows:

\begin{itemize}
\item We propose a novel filtering method based on PVI \cite{DBLP:conf/icml/EthayarajhCS22} to filter out examples that are not relevant or helpful to the desired intent.
\item We conduct experiments on three challenging intent detection datasets and show that our method achieves state-of-the-art  performance.
\item We conduct an in-depth study and present a comprehensive analysis of the factors that influence performance, including ablation studies and comparisons with alternative methods.
\end{itemize}

The rest of the paper is organized as follows: In \Cref{sec:related_work} we present relevant work and in \Cref{sec:method} we introduce our method. In sections 4 and 5 we discuss training details, experiments, and results. In section 6, we present our analysis and discuss alternative approaches we investigated. In section 7 we conclude, and in the following sections we discuss limitations and ethical considerations.

\section{Related Work}
\label{sec:related_work}
\paragraph{Intent Detection}
Intent detection is the task of identifying the user's intent by mapping the user's natural language utterance into one of several predefined classes \cite{DBLP:conf/naacl/HemphillGD90,DBLP:journals/corr/abs-1805-10190}.
It is a critical component in the pipeline of task-oriented dialogue systems, as it is used to determine the user's goal and to trigger an appropriate system action \cite{DBLP:conf/interspeech/RauxLBBE05,DBLP:journals/pieee/YoungGTW13}.
Several datasets have been proposed to evaluate the performance of intent detection models \cite[for some recent examples]{casanueva-etal-2020-efficient,DBLP:conf/iwsds/LiuESR19,larson-etal-2019-evaluation}. 
With the availability of such datasets, intent detection has been extensively studied in the literature.
Recently, pre-trained language models (e.g., BERT \cite{devlin-etal-2019-bert}) have been shown to be effective in intent detection \cite{DBLP:journals/corr/abs-2004-09936,zhang-etal-2020-discriminative,zhang-etal-2021-effectiveness-pre,zhang-etal-2021-shot,mehri-eric-2021-example}.

\paragraph{Data Augmentation}
Data augmentation is a widely-used technique to address the problem of data scarcity. 
Paraphrasing the data is one of the ways frequently used for augmentation and can produce more diverse synthetic text with different
word choices and sentence structures while preserving the meaning of the original text. Paraphrasing methods have been shown to be effective in many natural language processing tasks \cite{DBLP:conf/aaai/GuptaASR18,edunov-etal-2018-understanding,DBLP:conf/naacl/IyyerWGZ18,wei-zou-2019-eda,cai-etal-2020-data,okur-etal-2022-data,panda-etal-2021-multilingual,jolly-etal-2020-data}.
However, such methods often fail to generate more challenging and semantically diverse sentences that are important for the robustness of the downstream models.

Recently, conditional generation -- using a PLM to produce text conditioned on some label -- has become the dominant paradigm of data augmentation \cite{DBLP:conf/conll/BowmanVVDJB16,kumar-etal-2019-closer,DBLP:conf/aaai/Anaby-TavorCGKK20,DBLP:journals/corr/abs-2003-02245,yang-etal-2020-generative,DBLP:journals/corr/abs-2102-01335}. This is usually achieved by fine-tuning a language model to produce the original text given the label.

In the field of intent detection, previous work has proposed using data augmentation techniques to generate synthetic training data \cite{sahu-etal-2022-data,papangelis-etal-2021-generative}.
\citet{sahu-etal-2022-data} also used PLMs to generate augmented examples, but they require human effort for labeling.
This is a challenging task since it is expensive to annotate large amounts of data.

Our approach involves data valuation, similar to the concepts of \citet{DBLP:conf/icml/GhorbaniZ19,DBLP:conf/icml/MindermannBRS0X22}. However, our approach differs from such previous work in two key ways. First, \citet{DBLP:conf/icml/GhorbaniZ19} only evaluated the quality of the training set after training them, whereas we evaluate the synthetic examples before training the task model. Second, \citet{DBLP:conf/icml/MindermannBRS0X22} selected points that minimize the loss on a holdout set, whereas we select synthetic examples that are reasonably challenging to the task model. Our approach aims to address the problem of data scarcity by evaluating the synthetic examples generated by PLMs and selecting the most valuable examples to augment the training data.

\paragraph{In-context Learning}
Large language models such as GPT-3  \cite{DBLP:conf/nips/BrownMRSKDNSSAA20} and OPT \cite{DBLP:journals/corr/abs-2205-01068} have shown to be able to perform many natural language processing tasks with in-context learning. In this paradigm, the model is provided with a few exemplars based on which it performs the respective task.

In-context learning is a promising solution for few-shot learning.
Because of the effectiveness in few-shot performance, in-context learning has been applied to a wide range of NLP tasks. For dialogue tasks, in-context learning has been applied to intent classification \cite{yu-etal-2021-shot}, semantic parsing \cite{DBLP:conf/naacl/ShinD22}, and dialogue state tracking \cite{DBLP:journals/corr/abs-2203-08568}.

However, PLMs require a large amount of computational resources and the limitation on input length restricts the application of PLMs to intent detection tasks with large numbers of intents (e.g., 150 intents in CLINC \cite{larson-etal-2019-evaluation}), where we cannot fit examples for each intent in the input. One solution would be to call the model multiple times, each time with a subset of the possible intents. This would lead to increased inference time and may also impact performance.
Consequently, \citet{yoo-etal-2021-gpt3mix-leveraging,sahu-etal-2022-data} leveraged in-context learning and PLMs to generate synthetic examples for intent detection, instead of directly deploying the PLM.
However, they did not consider the quality of the generated examples, which may lead to the model overfitting on examples that are not relevant to the desired intent.


\section{In-Context Data Augmentation}
\label{sec:method}
In the following section, we describe our proposed two-stage method for data augmentation, which we refer to as In-Context Data Augmentation (\textbf{ICDA}).
The overall procedure is summarized in \mbox{Algorithm \ref{alg:icda}}. We apply ICDA to the task of few-shot intent detection, which involves classifying a user utterance $x$ into an intent label $y \in Y$. ICDA aims to generate synthetic examples $x^\prime$ such that they would belong to a given intent $y$.

\subsection{Synthesizing Examples}
\label{sec:synthetizing}

\begin{figure}[t]
    \centering
    \includegraphics[width=1.0\linewidth]{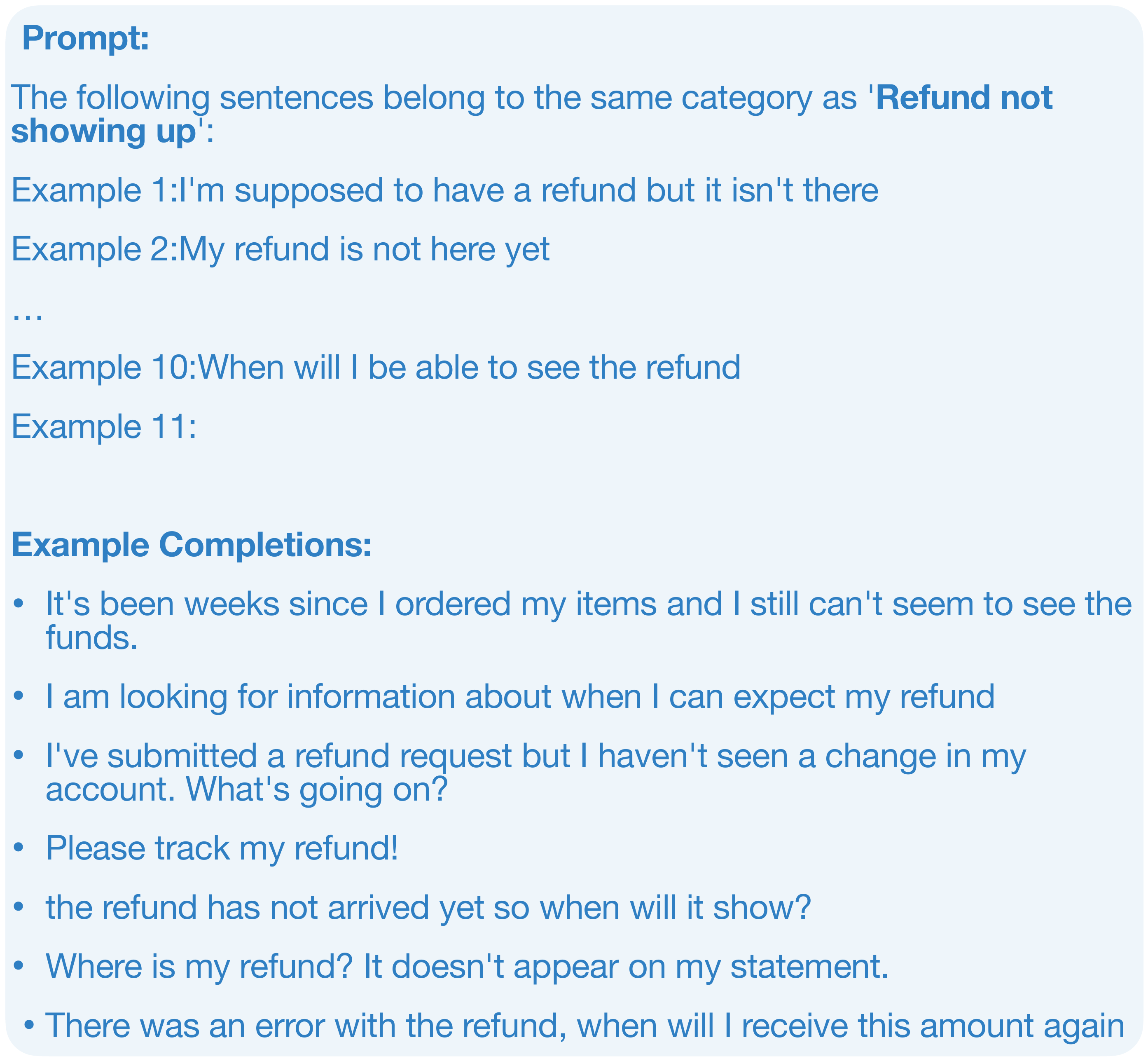}
    \caption{An example of the prompt used to generate synthetic examples.  The intent class is \textit{refund not showing up}. Completions  are generated by a pre-trained language model via sampling. Note that 5-shot experiments only use 5 examples from the training set.}
    \label{fig:synthetic_example}
\end{figure}

The core idea is to use a large pre-trained language model such as GPT-3 \cite{DBLP:conf/nips/BrownMRSKDNSSAA20} or OPT \cite{DBLP:journals/corr/abs-2205-01068} to generate synthetic data in the context of the training set.
In particular, for each intent class, we create a natural language context (prompt) that contains the intent class name, a set of real training examples under the same intent class, and an incomplete example.
For instance, the prompt for the intent class \textit{refund\_not\_showing\_up} is shown in Figure \ref{fig:synthetic_example}.
We feed the prompt to the language model and obtain a set of synthetic examples as outputs.
In this work, we use OPT-66B \cite{DBLP:journals/corr/abs-2205-01068}  as the language model to generate a set of examples for each intent class.
We adopt typical decoding
with $\tau = 0.9$ \cite{DBLP:journals/corr/abs-2202-00666} and set repetition penalty to $1.1$ following \citet{DBLP:journals/corr/abs-1909-05858} to generate the synthetic examples.\footnote{Implementation details are available from \url{https://huggingface.co/docs/transformers/main_classes/text_generation}}
Due to the fine-grained nature of  intents, and the sampling-based generation aiming to produce a set of diverse datapoints, we expect some of the generated utterances to not match the given intent.

Note that our method leverages PLMs in a way that is  orthogonal to the intent detection model.
Unlike other methods that use the same model to directly predict the intent class of a user utterance, we use a PLM to generate synthetic training instances. These instances are then used to augment the actual training data and train a smaller intent detection model.
This approach leverages the power of PLMs while preserving the independence of the intent detection model design.

\subsection{PVI Filtering}
\label{sec:filtering}
As mentioned above, given the stochastic nature of synthetic data generation, we expect some of the synthetic utterances not to match the given intent. To address this phenomenon, we filter generated instances and retain only those that are relevant and helpful to the desired intent classes.

Specifically, we apply Pointwise V-Information \cite{DBLP:conf/icml/EthayarajhCS22} - an idea originally suggested for understanding how difficult a dataset is - as a filter to discard unhelpful datapoints.
PVI of an utterance $x$ with respect to its corresponding intent class, $y$, is defined as:
\[
\mathrm{PVI}(x \to y) = -\log_2 g^* [\emptyset] (y) + \log_2 g' [x] (y)
\]
where, in this work, $g'$ and $g^*$ are the intent detection models finetuned with and without the input $x$, respectively. 
$\emptyset$ is a special token that is used to indicate the absence of an input utterance.

Intuitively, PVI measures the amount of information that the input $x$ provides to the intent detection model (compared to the absence of meaningful input).
A high PVI value indicates that the input $x$  provides a lot of information to the model, and thus is more likely to be helpful when training the model to classify instances of the intent class $y$. 
On the contrary, a low PVI value indicates that the input $x$ provides little information to the model, and thus is likely to be irrelevant to the intent \mbox{class $y$} \cite{DBLP:conf/icml/EthayarajhCS22}. 

We set a threshold $\epsilon$ (tunable parameter) to determine which $x$ are retained and conduct experiments to study the effect of the threshold in Section \ref{sec:analysis}. Algorithm \ref{alg:icda} defines $\epsilon$ as a function of $y$ to allow flexibility in its definition: either a fixed threshold for all intent classes, or a different threshold per intent class.

\begin{algorithm}[t]
\caption{In-Context Data Augmentation with PVI Filtering}
\label{alg:icda}
\DontPrintSemicolon
  \KwInput{Task Model $\mathcal{V}$, Language Model $\mathcal{PLM}$, Data Multiplier $m$, PVI Threshold Function $\epsilon$}
  \KwOutput{$\textnormal{Task Model } g$}
  \KwData{Seed Data $\mathcal{D}_{\textnormal{train}}=\{(\textnormal{input } x_i, \textnormal{gold label } y_i)\}^n_{i=1}$}
  
  $g' \gets \textnormal{Finetune } \mathcal{V} \textnormal{ on } \mathcal{D}_{\textnormal{train}} $
  
  $\emptyset \gets \textnormal{empty string}$

  $g^* \gets \textnormal{Finetune } \mathcal{V} \textnormal{ on } \{ (\emptyset, y_i) | (x_i, y_i) \in  \mathcal{D}_{\textnormal{train}} \}$
  
  $\mathcal{D}_{\textnormal{synthetic}} \gets \mathrm{Prompt}( \mathcal{PLM}, \mathcal{D}_{\textnormal{train}} )$

  \For{ $(x_i, y_i) \in \mathcal{D}_{\textnormal{synthetic}}$ }
  { 
    $\mathrm{PVI}(x_i \to y_i) \gets - \log_2 g^*[\emptyset](y_i) + \log_2 g'[x_i](y_i)$
  }
  
  $\mathcal{D}_{\textnormal{synthetic}} \gets
  \{
    (x_i, y_i) | (x_i, y_i) \in  \mathcal{D}_{\textnormal{synthetic}} 
    \And \mathrm{PVI}(x_i \to y_i) > \epsilon(y_i)
  \}$
  
  $g \gets \textnormal{Finetune } \mathcal{V} \textnormal{ on } \mathcal{D}_{\textnormal{train}} \bigcup \mathcal{D}_{\textnormal{synthetic} }$
\end{algorithm}

\section{Experimental Setup}
\label{sec:experiment}

\subsection{Datasets}
To evaluate the effectiveness of our approach in intent detection in cases where we have a large number of often semantically similar intent labels, we chose the BANKING \cite{casanueva-etal-2020-efficient}, HWU \cite{DBLP:conf/iwsds/LiuESR19}, and CLINC \cite{larson-etal-2019-evaluation} datasets and compare with recent state-of-the-art baselines.
BANKING comprises 13,083 utterances in a single banking domain and 77 intents. 
HWU includes 25,716 utterances with 64 intents across 21 domains. 
CLINC contains 23,700 utterances with 150 intents across 20 domains.

\subsection{Training}
In our experiments, we use RoBERTa-{\small LARGE} \cite{DBLP:journals/corr/abs-1907-11692} as the intent detection model $\mathcal{V}$ in Algorithm \ref{alg:icda}.
We use OPT-66B\footnote{We used p3dn.24xlarge AWS EC2 instances for our experiments.} \cite{DBLP:journals/corr/abs-2205-01068} as the language model $\mathcal{PLM}$ to generate synthetic examples and set the data multiplier $m$ to be $128$\footnote{This means that we generate $m$ times the available training data, e.g. (5 x 77) x $m$ in the 5-shot BANKING case.}.
We set the PVI threshold function $\epsilon$ to be the average PVI under each intent class in the validation set, where the PVI is computed using the same models as in Algorithm \ref{alg:icda}.
We train RoBERTa-{\small LARGE} for $40$ epochs with a batch size of $16$,  a learning rate of $1e-5$, and the AdamW optimizer \cite{DBLP:conf/iclr/LoshchilovH19}.
We use the HuggingFace Transformers library \cite{wolf-etal-2020-transformers} for all experiments. 

\begin{table}[t]
\centering
\small
\begin{tabular}{lcc}
\Xhline{2\arrayrulewidth}
& Full-shot mult. & Few-shot mult. \\
\hline
XS & - & 1x \\
S & 1x & 4x \\
M & 2x & 16x \\
L & 4x & 64x \\
XL & - & 128x \\
\Xhline{2\arrayrulewidth}
\end{tabular}
\caption{To assess the impact of the synthetic data size on performance, we experiment with several data multipliers (\emph{synthetic data size = source data size x mult.}).}
\label{tab:multipliers}
\end{table}

\subsection{Baseline Models}
\di{We need to clarify which baselines are for full-shot only and which ones are for both}

We compare our proposed method with the following baselines:

\noindent \textbf{RoBERTa-{\small BASE} + Classifier}  is a baseline that uses RoBERTa-{\small BASE} \cite{DBLP:journals/corr/abs-1907-11692}  with  a linear classifier on top \cite{zhang-etal-2020-discriminative}.

\noindent \textbf{USE} is a universal sentence encoder pre-trained on 16 languages supporting multiple down-stream tasks \cite{yang-etal-2020-multilingual}. 

\noindent \textbf{C{\small ONVE}RT} is an intent detection model finetuned from dual encoder models, which is pre-trained on (input, response) pairs from Reddit \cite{henderson-etal-2020-convert}.

\noindent \textbf{C{\small ONV}BERT} fine-tunes BERT on a large open-domain dialogue corpus with 700 million conversations \cite{DBLP:journals/corr/abs-2009-13570} .

\noindent \textbf{C{\small ONV}BERT + Combined} is an intent detection model based on C{\small ONV}BERT, with example-driven training based on similarity matching and observers for transformer attentions. It also conducts task-adaptive self-supervised learning with masked language modeling (MLM) on the intent detection datasets. Here, ``Combined" represents the best MLM+Example+Observers setting in the referenced paper \cite{mehri-eric-2021-example}.

\noindent \textbf{DNNC} (Discriminative Nearest-Neighbor Classification) is a discriminative nearest-neighbor model, which finds the best-matched example from the training set through similarity matching. The model conducts data augmentation during training and boosts performance by pre-training on three natural language inference tasks \cite{zhang-etal-2020-discriminative}.

\noindent \textbf{CPFT} (Contrastive Pre-training and Fine-Tuning) is the current state-of-the-art in few-shot intent detection on the selected datasets. It is pre-trained on multiple intent detection datasets in a self-supervised contrastive manner and then fine-tuned with supervised contrastive learning \cite{zhang-etal-2021-shot}.

\section{Experimental Results}
\label{sec:results}

\begin{table*}[t!]
\centering
\small
\begin{tabular}{lccccccccc}
\Xhline{2\arrayrulewidth}
                        & \multicolumn{3}{c}{\textbf{BANKING}} &  \multicolumn{3}{c}{\textbf{HWU}} &\multicolumn{3}{c}{\textbf{CLINC}}\\
                        Model & \textbf{5} & \textbf{10} & \textbf{Full} & \textbf{5} &  \textbf{10} & \textbf{Full}  & \textbf{5} & \textbf{10} & \textbf{Full} \\
                        \hline
                        RoBERTa-{\small Base} + Classifier 
                        & 74.04 & 84.27 & - & 75.56 & 82.90 & - & 87.99 & 91.55 & -\\
                        USE 
                        &  76.29 & 84.23 & 92.81  & 77.79 & 83.75 & 91.25 & 87.82 & 90.85 & 95.06 \\
                        C{\small ONVE}RT 
                        & 75.32 & 83.32 & 93.01 & 76.95 & 82.65 & 91.24 & 89.22 & 92.62 & 97.16 \\
                        USE+C{\small ONVE}RT 
                        & 77.75 & 85.19 & 93.36 & 80.01 & 85.83 & 92.62 & 90.49 & 93.26 & 97.16\\
                        C{\small ONV}BERT 
                        & - & 83.63 & 92.95 & - & 83.77 & 90.43 & - & 92.10 & 97.07 \\
                        \quad + MLM 
                        & - & 83.99 & 93.44 & - & 84.52 & 92.38 & - & 92.75 & 97.11 \\
                        \quad + MLM + Example
                        & - & 84.09 & 94.06 & - & 83.44 & 92.47 & - & 92.35 & 97.11 \\
                        \quad + Combined 
                        & - &  85.95  & 93.83 & - & 86.28 & {\bf 93.03} & - & 93.97 & {\bf 97.31} \\
                        DNNC 
                        & 80.40 & 86.71 & - & 80.46 & 84.72 & - & 91.02 & 93.76 & - \\
                        CPFT 
                        & 80.86 & 87.20 & - & 82.03 & 87.13 & - & 92.34 & 94.18 & - \\
                        \hline
                        RoBERTa-{\small Large} + Classifier
                        & 78.99 & 86.08 & 93.70 & 74.44 &  84.11 & 92.13 & 89.89 & 93.56 & 96.80 \\
                        \quad + \textbf{ICDA}-XS  & 80.29 & 86.72 & - & 81.32 & 85.59 & - & 91.16 & 93.71 & - \\
                        \quad + \textbf{ICDA}-S  & 81.95 & 87.37 & 93.66 & 81.97 & 86.25 & 92.33 & 91.22 & 93.98 & 96.97 \\
                        \quad + \textbf{ICDA}-M & \textbf{84.01$^*$} & 88.64 & 93.73 & 81.84 & 87.36 & 92.12 & 91.93 & 94.71 & 97.06 \\
                        \quad + \textbf{ICDA}-L & 83.90 & 89.12 & \textbf{94.42$^*$} & 81.97 &  86.94 & 92.57 & 92.41 & 94.73 & 97.12 \\
                        \quad + \textbf{ICDA}-XL & 83.90 & \textbf{89.79$^*$} & - & \textbf{82.45$^*$} & \textbf{87.41$^*$} & - & \textbf{92.62$^*$} & \textbf{94.84$^*$} & -\\
\Xhline{2\arrayrulewidth}
\end{tabular}
\caption{Intent Detection Accuracy (in \%) in few-/full-shot settings with augmented data from OPT-66B. Numbers in bold are the best results and numbers with $^*$ are statistically significant by t-test  ($p<0.05$) compared to the baselines (\textbf{5} / \textbf{10} examples per intent).}
\label{tab:intent_results}
\end{table*}

We conduct experiments on three benchmark datasets to validate the effectiveness of our proposed method. 
We first use OPT-66B to generate augmentation examples and then apply our method to enhance a RoBERTa-Large model trained on three datasets. 
We repeat all experiments with 5 random seeds and report the average performance in Full-shot and Few-shot settings. To investigate the effect of the synthetic data size, we experiment with a variety of multipliers (see Table \ref{tab:multipliers} for notations).
Results are shown in Table \ref{tab:intent_results}.

\textbf{Full-shot settings.}
In this setting, we use the entire training set for each domain.
The proposed method  achieves the best performance on BANKING and comparable results on HWU and CLINC.
In particular, on BANKING, we improve the C{\small ONV}BERT + Combined baseline \cite{mehri-eric-2021-example} by 0.59\% (absolute) and the RoBERTa-Large baseline by 0.72\% (absolute).
Compared with the C{\small ONV}BERT + Combined, which is pre-trained on intent detection datasets in a self-supervised fashion and adds examples-driven training and specific model architectural design, our method achieves similar results with much simpler model design. 
Furthermore, our method is orthogonal to model architectures and can be integrated with any other approach for further improvement.

We also find that ICDA improves the performance of the RoBERTa-Large model on HWU and CLINC.
This highlights the effectiveness of our method for enhancing intent detection models.
Moreover, state-of-the-art performance on BANKING with the proposed method and RoBERTa-Large shows that our method is capable of generating high-quality augmentation examples to enhance the RoBERTa-Large model on the most fine-grained intent detection task. 


\textbf{Few-shot settings.} In this setting we only use a small number of instances (datapoints) per class.
We evaluate our method in both 5-shot and 10-shot settings and compare it with several strong baselines.
Our proposed method outperforms all baselines on all datasets in both 5-shot and 10-shot settings.
ICDA-M achieves the best performance in 5-shot settings on BANKING dataset and ICDA-XL achieves the best performance on HWU and CLINC datasets in 5-shot settings and on all datasets in 10-shot settings.
All configurations of our method significantly improve the performance of a RoBERTa-Large model trained on any of the three datasets.
Compared with CPFT \cite{zhang-etal-2021-shot}, which utilizes contrastive learning for few-shot intent detection with extra data, our method achieves better performance without any additional human-annotated data.
This showcases the advantage of our method for few-shot intent detection.

We also observe that our method consistently improves the performance of the baseline model as the number of synthetic datapoints increases from XS to XL.
This indicates that the generated instances from our method can gradually cover more and more information of real instances and are capable of providing more useful information for model training. 


\begin{table}[t]
\centering
\small
\begin{tabular}{llrrr}
\Xhline{2\arrayrulewidth}
                       & Model & BANKING & HWU & CLINC\\
                       \hline
                        & RoBERTa-{\small Large} 
                        & 86.08 & 84.11 & 93.56\\
                        \hline
                        & All & 84.19 & 84.57 & 94.24 \\
                        & All w/ relabeling & 87.05 & 85.22 & 93.02\\
                        \hline
                        \multirow{4}{*}{{\rotatebox[origin=c]{90}{ {\small PVI} }}} & Global Low PVI & 73.99  & 69.61 & 85.42 \\
                        & Global High PVI & 87.38 & 86.27 & 94.27 \\
                        & Per-Intent Low PVI & 76.49 & 71.84 & 89.33 \\
                        & Per-Intent High PVI & \textbf{88.64} & \textbf{87.36} & \textbf{94.71} \\
\Xhline{2\arrayrulewidth}
\end{tabular}
\caption{Intent Detection Accuracy (in \%) for RoBERTa-Large model in 10-shot settings with ICDA-M synthetic instances from OPT-66B. Numbers in bold are statistically significant by t-test ($p<0.05$). ``All'' represents using all synthetic data without PVI filtering. and ``All w/ relabeling" represents using ``All" and an oracle intent classifier to relabel the synthetic data.}
\label{tab:threshold}
\end{table}

\section{Analysis and Discussion}
\label{sec:analysis}

In this section, we analyze the performance of ICDA and other approaches we tried. 
We first identify several factors that affect performance, and then present evidence that ICDA works by transferring knowledge from the pretrained generator to the task model. We then discuss a data-relabelling experiment and an experiment using uncertainty measures or data cartography \cite{swayamdipta2020dataset} as filters.

\subsection{Factors that Affect ICDA Performance}

\noindent \textbf{ICDA is effective at various training sizes.}
Throughout this work, we conduct experiments with different seed data sizes\footnote{By \emph{seed data}, we mean data taken from each dataset, i.e. not synthetic data produced by ICDA.} to study the effect of training size.
By looking at the results in Table \ref{tab:intent_results}, we observe that our proposed method consistently improves the accuracy of the downstream model in all training sizes.
Also, as the training size decreases, we see that the ICDA improvement increases significantly.
For example, on BANKING, the improvement goes from $0.72\%$ in the full shot setting to $5.02\%$ as the training size decreases to 5-shot.
This indicates that ICDA is more effective when we have few training data available. \\

\noindent \textbf{PVI filtering threshold.}
To study the effect of the threshold function $\epsilon$, we conduct experiments with two different threshold functions: \textit{Global}, and \textit{Per-Intent}. 
\textit{Global} means that the PVI threshold is the same for all intent classes, which is the average PVI value in the validation set.
\textit{Per-Intent} means that the PVI threshold is different for each intent class, which is the average PVI value under each intent class in the validation set. As a sanity check, we also conduct experiments using synthetic instances with PVI values lower than the threshold (\textit{Low PVI)} as opposed to the normal (\textit{High PVI)} instances.

We show the results in Table \ref{tab:threshold} (bottom half), where we see that \textit{Per-Intent High PVI} filtering performs the best. 
Compared to using all synthetic training data without filtering (referred to as \textit{All}), we see that \textit{High PVI} filtering in general helps in improving accuracy.
In BANKING, for example, when PVI filtering is applied with \textit{Per-Intent High PVI}, the accuracy is $88.64\%$ with 10-shot training size, which is significantly better than the result without PVI filtering ($84.19\%$) -- the same holds for the other two datasets. 
For the \textit{Low PVI} conditions, we observe that performance drops significantly.
This indicates that the model overfits on those examples that are not relevant to the desired intent. We discuss the \textit{All w/ relabelling} condition in Section 6.3. 

\begin{figure}[ht]
    \centering
    \includegraphics[width=1\linewidth]{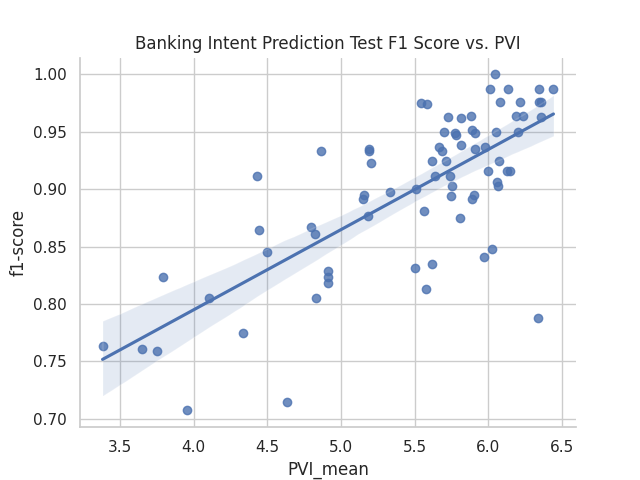}
    \caption{Intent Detection F1 score per intent class (circle) of the BANKING test set, justifying why we need a PVI threshold per intent.}
    \label{fig:f1_pvi}
\end{figure}

In Figure \ref{fig:f1_pvi}, we plot the F1 score against the PVI score of the test set instances grouped by intent, showing that some classes are harder than others, further supporting why we need a threshold per class rather than a global one.

\begin{table*}[h!]
\small
\centering
\begin{tabular}{l l p{0.6\linewidth} r}
						Data & Prompt Label & Generated Sentence & PVI \\ 
                        \hline
                        BANKING & Refund not showing up & I didn't see my refund appear on my statement.$\dagger$ & \textbf{6.10} \\ 
                         & (PVI Threshold: 5.79) & Where did your refund end up at? Please send it back immediately since I had a return and then refunded your purchase in full and I want it all returned back to my credit card.$\dagger$ & \textbf{5.81} \\ 
                        & & Can we please cancel my return \& resend my goods again & 3.97 \\ 
                        & & Please confirm what is the reason for delay in payment for my purchase?	& -3.86 \\ 
                        \hline
                        HWU & alarm query & show me all the alarms and remind me about them$\dagger$	& \textbf{5.69} \\ 
                        & (PVI Threshold: 4.28) & i want to be notified when alarm goes off.$\dagger$ & \textbf{4.36} \\ 
                        & & how do i delete or disable alarms. & 3.18 \\ 
                        & & list all the events on this date & -5.13 \\ 
                        \hline
                        CLINC & accept reservation & does hanover steakhouse take reservations$\dagger$ & \textbf{6.74} \\ 
                        & (PVI Threshold: 6.53) & are there any restaurants that take reservations for dinner in philadelphia$\dagger$	& \textbf{6.58} \\ 
                        & & how many days prior is required for making reservations & $5.39$ \\ 
                        & & what time does bibiana's in greenwich open & $-4.31$ \\ 

\end{tabular}
\caption{Synthetic examples generated from OPT-66B. $\dagger$ indicates the sentences that belong to the same intent as the prompt label from our manual assessment; and bold denotes the PVI values over the threshold for given label.}
\label{tab:qualitative_example}
\end{table*}

\subsection{Why Does ICDA Work?}

\noindent \textbf{PVI filtering discards mislabeled examples.}
We believe that the success of ICDA is because of not only  the high diversity of the synthetic instances produced by the generator, but also  the fact that PVI filtering effectively discards digressed instances.
To verify this hypothesis, we randomly sample several synthetic instances from the OPT-66B generator and manually assess if each instance follows the same intent as the prompt label.
We show some examples in Table \ref{tab:qualitative_example}.\di{Besides manual evaluating several examples, I think we need to perform such human eval on a larger size of samples, e.g. 200.}
We observe that instances that are relevant to the desired intent are assigned high PVI values, and instances that are not relevant to the desired intent are assigned low PVI values.
This further indicates that the per-intent threshold function provides an effective indicator of relevance.
For example, in the BANKING dataset, most relevant instances have PVI values greater than $5.79$, and most non-relevant instances have PVI values less than $5.79$.
This indicates that PVI filtering is an effective method for discarding mislabeled data points. \\

\noindent \textbf{ICDA produces fluent and diverse utterances.}
We hypothesize that our proposed method is effective because it introduces more fluent and diverse utterances. We therefore compare synthetic data under the 10-shot XS condition (i.e., we generate 10 synthetic datapoints) with the original 10-shot datapoints taken from the training data.
Then we use a GPT2 model trained on the test set of each benchmark dataset to calculate the perplexity of the generated utterances.
We also use the same synthetic set to calculate the distinct-1, distinct-2, self-BLEU, and perplexity (PPL) metrics.
We report the results in Table \ref{tab:distint_ppl} and observe that our proposed method generates more diverse utterances as shown by distinct-1, distinct-2, and self-BLEU.
This indicates that our proposed method harnesses the generation power of the OPT-66B generator.
Additionally, the perplexity of synthetic utterances is slightly higher than the human-annotated training set.
These results suggest that our proposed method generates more diverse utterances, which can help the task model to learn a better representation.

\begin{table}[]
\centering
\small
\begin{tabular}{llcccc}
\toprule
& & & & Self- & \\
Data & Split & D-1 ↑ & D-2 ↑ & BLEU ↓ & PPL ↓ \\
\midrule
\multirow{3}{*}{Bank.} & Test          & -        & -        & -           & 12.14 \\
\cline{2-6}
		& 10-shot 
		& 0.15   & 0.54   & 0.24      & \textbf{17.34} \\
        & ICDA 
        & \textbf{0.21}   & \textbf{0.66}   & \textbf{0.11}      & 21.33 \\
        \hline
\multirow{3}{*}{HWU}     & Test          & -        & -        & -           & 14.84 \\
\cline{2-6}
        & 10-shot 
        & 0.25   & 0.71   & 0.07      & \textbf{26.97} \\
        & ICDA 
        & \textbf{0.30}   & \textbf{0.78}   & \textbf{0.03}      & 28.52 \\
        \hline
\multirow{3}{*}{CLINC}   & Test          & -        & -        & -           & 14.77 \\
\cline{2-6}
		& 10-shot 
		& 0.15   & 0.49   & 0.28      & \textbf{34.23} \\
        & ICDA 
        & \textbf{0.20}   & \textbf{0.60}   & \textbf{0.17}      & 37.34\\
\bottomrule
\end{tabular}
\caption{Quantitative metrics of fluency and diversity of real and synthetic utterances in 10-shot settings as measured with distinct-1 (D-1), distinct-2 (D-2), self-BLEU, and perplexity.}
\label{tab:distint_ppl}
\end{table}





\subsection{Data Relabelling}
Following \citet{sahu-etal-2022-data}, we wanted to see if it is effective to use the available data to train an intent classifier and then use it to relabel the synthetic data. Intuitively, such a method would correct mistakes in the generation process. To test the feasibility of this approach, we train an oracle classifier using the entire training data of each dataset and use this as an upper bound. The results are shown in Table \ref{tab:threshold} (``All w/ relabeling"), where we see that while promising, this approach underperforms ICDA.

\section{Conclusion}
We introduced \textbf{I}n-\textbf{C}ontext \textbf{D}ata \textbf{A}ugmentation, a novel data augmentation framework to generate synthetic training data, preserving quality and diversity.
We demonstrate that ICDA is effective on multiple intent detection benchmarks, with state-of-the-art few-shot performance.
Our analysis shows that ICDA tends to perform better in low-resource settings and that our PVI filtering strategy is important for performance.
Future work includes applying ICDA to other conversational understanding tasks such as slot filling and dialogue state tracking, and incorporating other filtering or data selection strategies for further performance gains.

\section*{Limitations}
\begin{figure}[ht]
    \centering
    \includegraphics[width=1\linewidth]{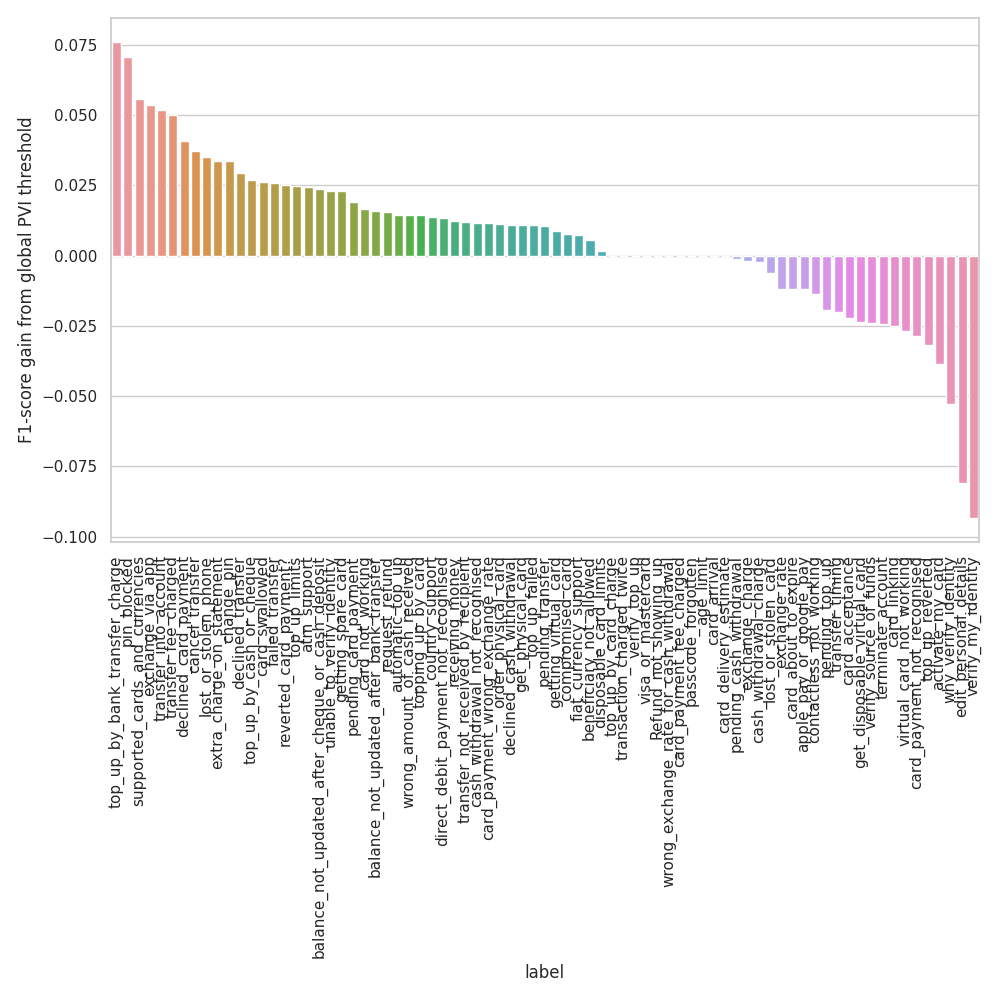}
    \caption{This figure shows the difference in Intent Detection F1 score for each intent, if we have a PVI threshold per-class VS having a fixed PVI threshold. See larger figure in Appendix.}
    \label{fig:local_vs_global}
\end{figure}

In this section we take BANKING as a case study to motivate PVI and discuss some of the limitations of our approach. Figure \ref{fig:local_vs_global} shows how much we gain (or lose) in F1 score when we use a custom threshold for each class vs. a fixed threshold. While most classes benefit, there are clearly many that show performance degradation. Another limitation is the size of the model we use to generate synthetic instances (OPT-66B); in general the larger the model is, the better the generated data is.

\section*{Ethical Considerations}
As with any work involving PLMs (or \emph{foundation models}), due to the data and training methods, there is inherent risk of generating biased, toxic, harmful, or otherwise unwanted output. Regarding our work in particular, as we show in Figure \ref{fig:local_vs_global}, the model's performance on some of the classes can degrade. More analysis needs to be done before deploying our approach, since it is unclear whether it will introduce a bias towards certain types of classes.

\bibliography{anthology,custom}
\bibliographystyle{acl_natbib}

\appendix

\section{Data Cartography and Uncertainty}
Apart from relabelling, we investigated two additional approaches to rank synthetic instances as easy or hard to classify. We used data cartography \cite{swayamdipta2020dataset} and classification uncertainty to guide our filtering. 

Data cartography classifies the training data in four categories:  Easy-to-learn, Low-Correctness, Ambiguous, Hard-to-Learn using training dynamics (i.e. the model’s confidence in the true class, and the variability of this confidence across epochs).

For uncertainty modeling, we assign uncertainty scores to each training instance in a \textit{cross-validation} manner. 
We first split the training set into 5 folds, hold one fold out as validation, and predict on the validation with the classifier trained on the remaining 4 folds.
We tried the following uncertainty measures: Contrastive Active Learning (AL) \cite{margatina-etal-2021-active}, Least Confidence \cite{DBLP:conf/aaai/CulottaM05}, Prediction Entropy \cite{DBLP:conf/icml/SchohnC00,DBLP:conf/icml/RoyM01}, and Breaking Ties \cite{DBLP:conf/ida/SchefferDW01,DBLP:conf/icpr/LuoKGHSRH04}. 

We conducted experiments using the above approaches to select data that amounts to one third of the total training data in BANKING (i.e., we select the top 33\% hardest examples, etc.). As an additional baseline, we include a random filter, i.e., a randomly sampled 33\% portion of BANKING. Table \ref{tab:cartography} shows the results, where we see that the performance actually degrades when compared to using the entirety of the data. We experimented with a few more variations in the filtering thresholds but no combination improved performance and we do not report those results here. See Figures \ref{fig:cartography_10shot} and \ref{fig:cartography_fullshot} in the Appendix~\ref{sec:appendix_figure} for a visualization of the BANKING data map.

\begin{table}[t]
\centering
\small
\begin{tabular}{llrr}
\Xhline{2\arrayrulewidth}
    & & 100\% Train & 92.89 \\
    \hline
    \multirow{13}{*}{{\rotatebox[origin=c]{90}{ {\small 33\% Train} }}} & & Random & 89.50\\ 
    \cline{2-4}
    & \multirow{6}{*}{{\rotatebox[origin=c]{90}{ {\small Uncertainty} }}} & & \\
    & & Contrastive AL & 88.54 \\
    & & Least Confidence & 89.08 \\
    & & Breaking Ties & 89.20 \\
    & & Prediction Entropy & 89.23 \\
    & & & \\
    \cline{2-4}
    & \multirow{6}{*}{{\rotatebox[origin=c]{90}{ {\small Cartography} }}} & & \\
    & & Easy to Learn & 90.44 \\
    & & Ambiguous & 90.94 \\
    & & Low Correctness & 91.00 \\
    & & Hard to Learn & 91.26 \\
    & & & \\
\Xhline{2\arrayrulewidth}
\end{tabular}
\caption{Intent Detection Accuracy (in \%) for ConvBERT model, trained on different selections of BANKING77 under full-shot settings.}
\label{tab:cartography}
\end{table}

\section{Figures}
\label{sec:appendix_figure}

\begin{figure*}[ht]
    \centering
    \includegraphics[width=1\linewidth]{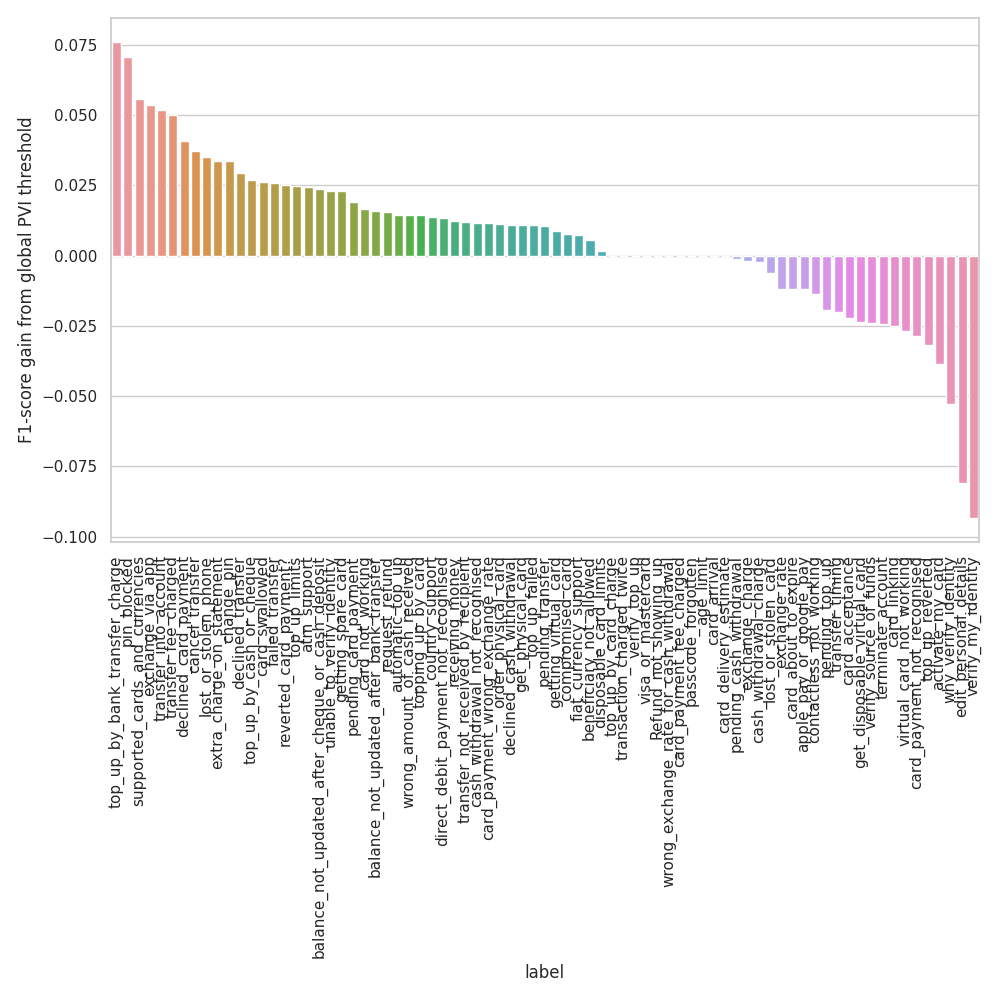}
    \caption{This figure shows the difference in F1 score for each intent, if we have a PVI threshold per-class VS having a fixed PVI threshold (Enlarged Figure \ref{fig:local_vs_global}).}
    \label{fig:local_vs_global_large}
\end{figure*}

\begin{figure*}[ht]
    \centering
    \includegraphics[width=1\linewidth]{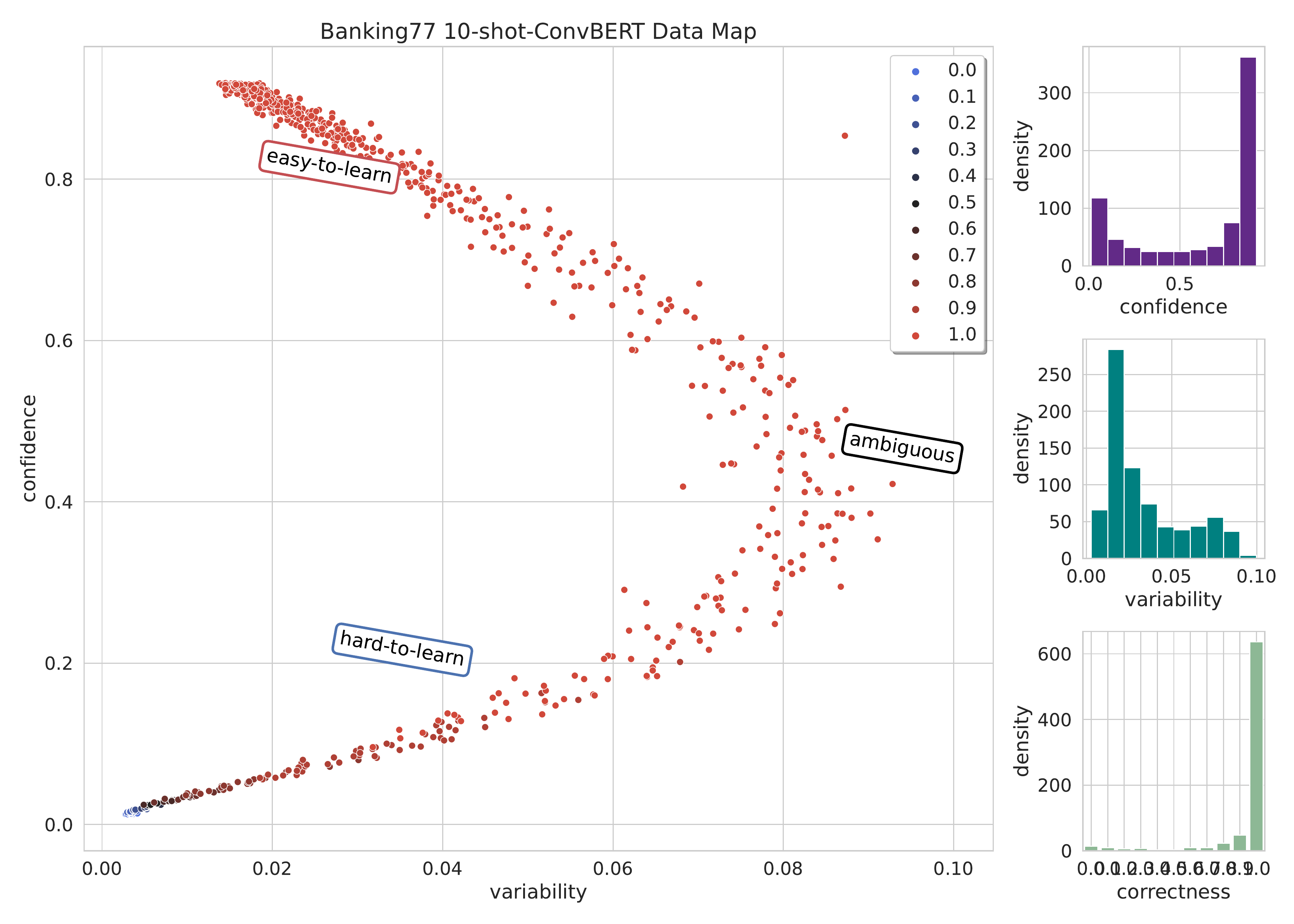}
    \caption{Data map for BANKING (10-shot).}
    \label{fig:cartography_10shot}
\end{figure*}

\begin{figure*}[ht]
    \centering
    \includegraphics[width=1\linewidth]{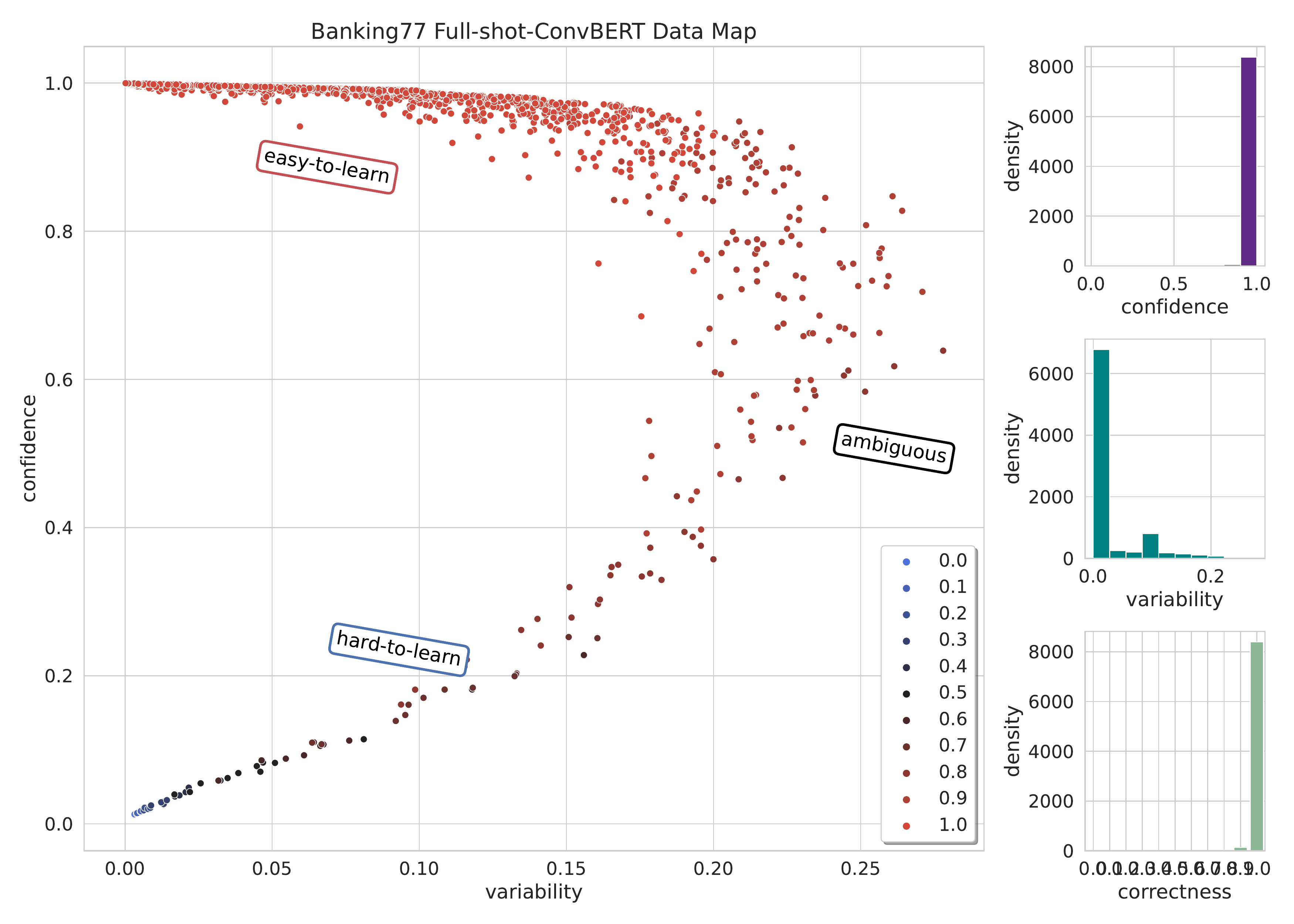}
    \caption{Data map for BANKING (full-shot).}
    \label{fig:cartography_fullshot}
\end{figure*}

\end{document}